\documentclass[sigconf]{acmart}

\usepackage{booktabs} 

\setcopyright{rightsretained}

\usepackage{booktabs} 
\usepackage{subfig}
\usepackage{epsfig}

\begin{document}
\title[Adaptive Discriminative Region Discovery for Scene Recognition]{From Volcano to Toyshop: Adaptive Discriminative Region Discovery for Scene Recognition}

\author{Zhengyu Zhao}
\affiliation{Radboud University, Netherlands}
\email{z.zhao@cs.ru.nl}

\author{Martha Larson}
\affiliation{Radboud University and TU Delft, Netherlands}
\email{m.larson@cs.ru.nl}


 \renewcommand{\shortauthors}{}

\begin{abstract}
  
As deep learning approaches to scene recognition emerge, they have continued to leverage discriminative regions at multiple scales, building on practices established by conventional image classification research. 
However, approaches remain largely generic, and do not carefully consider the special properties of scenes. 
In this paper, inspired by the intuitive differences between scenes and objects, we propose Adi-Red, an adaptive approach to discriminative region discovery for scene recognition. 
Adi-Red uses a CNN classifier, which was pre-trained using only image-level scene labels, to discover discriminative image regions directly. 
These regions are then used as a source of features to perform scene recognition. 
The use of the CNN classifier makes it possible to adapt the number of discriminative regions per image using a simple, yet elegant, threshold, at relatively low computational cost. 
Experimental results on the scene recognition benchmark dataset SUN397 demonstrate the ability of Adi-Red to outperform the state of the art. 
Additional experimental analysis on the Places dataset reveals the advantages of Adi-Red, and highlight how they are specific to scenes. 
We attribute the effectiveness of Adi-Red to the ability of adaptive region discovery to avoid introducing noise, while also not missing out on important information.

\end{abstract}



\copyrightyear{2018} 
\acmYear{2018} 
\setcopyright{acmlicensed}
\acmConference[MM '18]{2018 ACM Multimedia Conference}{October 22--26, 2018}{Seoul, Republic of Korea}
\acmBooktitle{2018 ACM Multimedia Conference (MM '18), October 22--26, 2018, Seoul, Republic of Korea}
\acmPrice{15.00}
\acmDOI{10.1145/3240508.3240698}
\acmISBN{978-1-4503-5665-7/18/10}

\maketitle

\begin{figure}[tbp]
\centering
\subfloat[campsite]{\includegraphics[height=0.4\columnwidth,width=0.9\columnwidth]{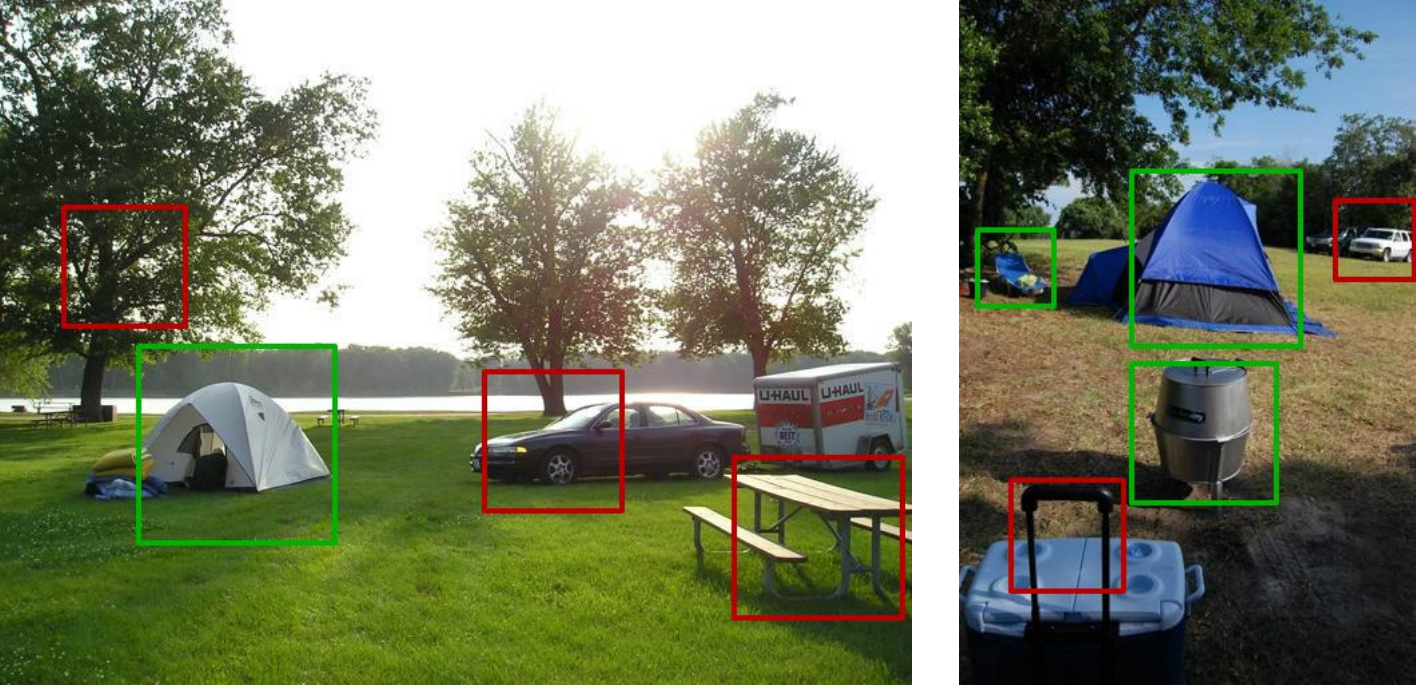}}
\hspace{0.2cm}\\
\subfloat[art\_school]{\includegraphics[height=0.4\columnwidth,width=0.9\columnwidth]{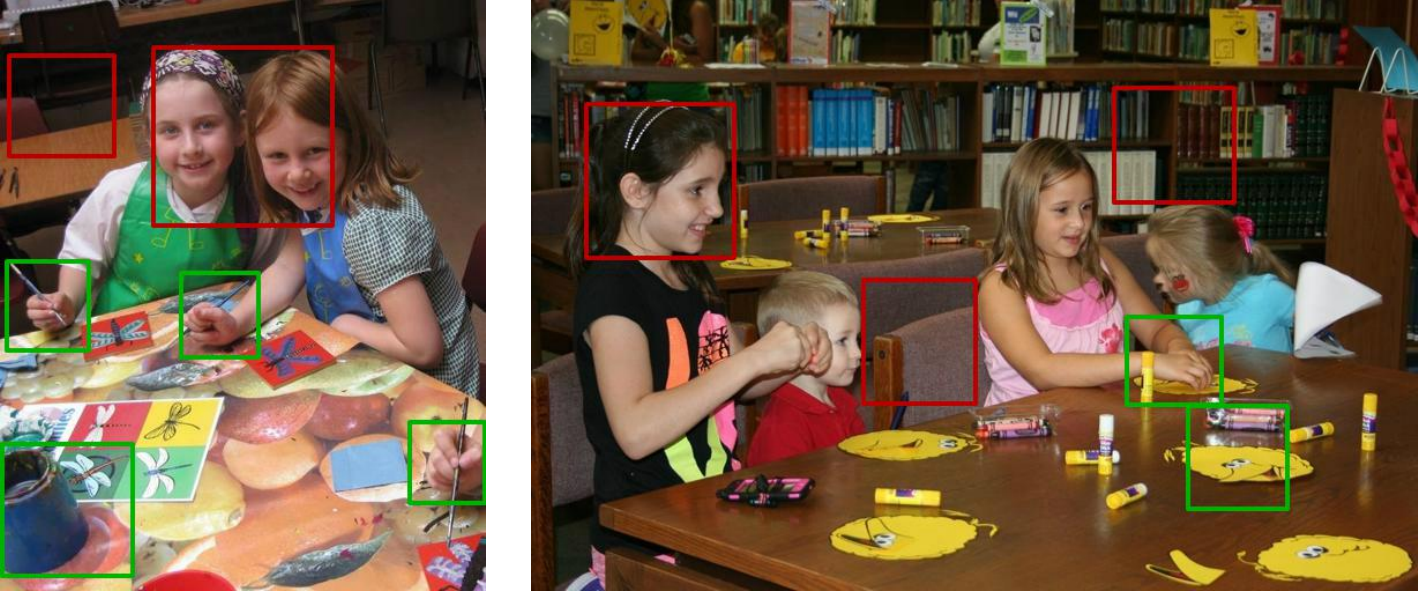}}
\caption{Examples from two scene categories in the Places dataset. The squares are characteristic regions for the scene category. Green outlined squares are highly discriminative: their number varies across images. [Best viewed in color]
}
\label{First_figure}
\end{figure}
\section{Introduction}
\label{introduction}
Deep learning approaches achieve impressive performance in scene recognition. 
However, the full potential of deep learning for scene recognition does not lie in generic Convolutional Neural Network (CNN) approaches, but rather in approaches that seek performance improvements based on the properties that make scenes special.  
In this paper, the basic conceptual differences between scene recognition and object recognition motivate an approach that can achieve improvement in scene recognition, while at the same time improving computational efficiency.\footnote{Our code is available at https://github.com/ZhengyuZhao/Adi-Red-Scene} 

The basic conceptual differences important for our approach are inherent in our intuitive understanding of ``object'' vs. ``scene''.
An object consists of contiguous parts. 
An image of an object can contain background regions that are clearly not part of the object. 
In contrast, the presence, absence, or relative configuration of elements of a scene is not particularly well constrained. 
An image of a scene can contain a background, however, that background is part of the scene.
In short, considered conceptually, the nature of scenes leads us to an approach that, in contrast to a generic approach, minimizes assumptions about the number and the position of the regions of an image that are important for scene classification.

The bridge between the conceptual properties of scenes and the algorithm that we propose in this paper is illustrated by examples from two scene categories in Fig. 1. 
The boxes in each image indicate image regions that are representative for the scene category. 
We see that multiple regions from various parts of the image serve to characterize the scene as being a member of a scene category. 
Further, we see that not all characteristic regions are equally useful to discriminate scene categories. 
Intuitively, regions with the green outline are highly discriminative, and the regions with the red outline are characteristic, but not highly discriminative. 
The key insight and novelty of our approach is to use a deep learning classifier to identify discriminative regions in a way that makes it possible to vary the number of regions per image with a simple, yet elegant, threshold.

Our idea represents an important extension to the existing work on scene recognition, which we describe briefly here, and discuss in more depth in the related work in Section~\ref{related}.
We build on the longstanding recognition that features extracted from discriminative regions at multiple scales improve scene recognition. This insight predates the deep learning era. 
Specifically, ~\cite{singh2012unsupervised,juneja2013blocks,lin2014learning} discovered discriminative regions by iteratively learning patch-based detectors on the dense image patches from the training set.
This work applied a sliding-window at multiple scales and constructed the final image-level descriptors by summarizing all the detection responses with spatial encoding methods such as Spatial Pyramid Matching (SPM)~\cite{lazebnik2006beyond}. 
These approaches made use of hand-crafted features (i.e., HOG~\cite{dalal2005histograms}).
They also suffered scalability issues, since iterative learning was applied to a huge number of multi-scale patches. 

When discriminative regions started to be used in deep-learning based approaches to scene recognition, they tried to move away from using a large number of patches per image, but they have continued to use a fixed number of regions per image.
Typically, recent methods~\cite{wu2015harvesting,xie2017hybrid} have relied on region proposal techniques from the object recognition literature~\cite{uijlings2013selective,arbelaez2014multiscale} to learn a collection of discriminative filters for each category. 
If we consider the conceptual properties of scene images discussed above, we can project approaches that impose assumptions about the number of regions per image will miss useful information or introduce noise. 

Our proposed approach, adaptive discriminative region discovery (Adi-Red), addresses the problem of using an inflexible number of discriminative image regions to extract features for scene recognition.
Adi-Red uses a class activation map from a CNN classifier~\cite{zhou2016learning} in order to identify multiple discriminative region candidates.
To our knowledge Adi-Red is the first approach to deploy such a CNN to discover discriminative regions for scene recognition.
Because the candidate set is broad, we minimize constraints on the positions of the discriminative regions. 
Multiple positions across the image are possible.
We then isolate only the most useful discriminative regions by using a threshold, which allows the number of regions used per image for scene recognition to vary. 
The title of the paper includes the phrase `from volcano to toyshop' since, as we will see in the experimental results, Adi-Red is able to automatically address the difference between the two. 
Specifically, Adi-Red selects a different number of discriminative regions for images of scenes like `volcano', which contain a few discriminative regions, and for images of scenes like `toyshop', which contain a larger number.

Two additional aspects of Adi-Red are important.
First, current deep learning approaches ~\cite{wu2015harvesting,xie2017hybrid,cheng2018scene} that use discriminative regions all still suffer from the computational load of laborious clustering and screening operations on a large number of patches.
Adi-Red derives information about discriminative regions directly from the CNN, and eliminates this need.
Second, recent top-performing approaches, i.e.,~\cite{cheng2018scene} to region selection for scene classification use the appearance of objects to select patches.
By deriving information directly from the CNN, Adi-Red can be guided by objects, but is also free to take other visual patterns into account.

The key contributions of this paper are summarized as:
\begin{itemize} 
\item We propose Adi-Red, a classifier-based region discovery approach that derives discriminative information directly from a CNN classifier, which was pre-trained on only image-level scene labels (i.e., no object detection or object-level labels are needed).
\item We demonstrate the strength of using an adaptive number of regions per image to extract features capturing local-level information. This insight is based on an intuitive conceptual understanding of scenes, but has not previously been applied in scene recognition.
\item Adi-Red improves efficiency by eliminating the need for both feature extraction from a large number of local patches and clustering operations, which are common in current methods.
\end{itemize}

The paper is organized as follows. In Section~\ref{related}, we cover related work that introduces scene recognition and highlights how our approach extends the state of the art. The Adi-Red approach is introduced in Section~\ref{approach}. Section~\ref{experiments} describes the experimental set up and results, and discusses the implications of our results in details. Finally, Section~\ref{conclusions} summarizes and provides an outlook on future work.

\begin{figure*}
\centering
\includegraphics[height=0.85\columnwidth,width=0.9\textwidth]{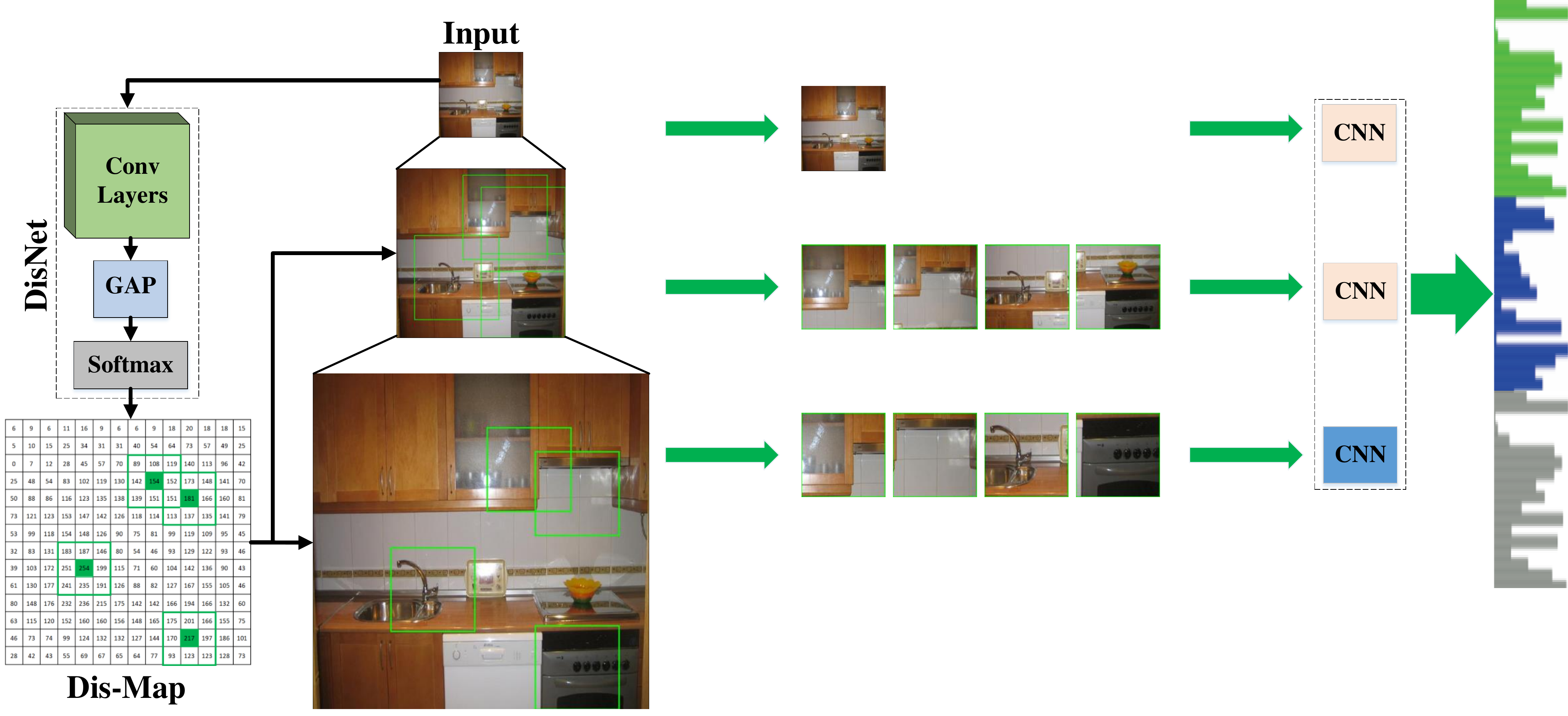}

\caption{The framework of our adaptive discriminative region discovery (Adi-Red) approach. A discriminative discovery network (DisNet) generates discriminative map (Dis-Map) for the input image. A threshold $T$ on the local maxima of Dis-Map is further applied to select the discriminative locations (green squares). The corresponding image regions from two different scales of input image are used to complement the global-level image by adopting a multi-scale pipeline for CNN feature extraction.
}
\label{diagram}
\end{figure*}

\section{Related work}
\label{related}
In this section, we will briefly review the history of scene recognition, and specifically present the differences between related work and our approach.

\subsection{Scene Recognition}
Scene recognition has a long interesting history. To address the complex characteristics of scenes, early work extracted local descriptors (SIFT~\cite{lowe2004distinctive}, HOG~\cite{dalal2005histograms}) to complement global, image-level features (Gist~\cite{oliva2001modeling}). 
As mentioned in the introduction, the importance of discriminative regions predates deep learning.
Specifically,~\cite{singh2012unsupervised,juneja2013blocks,lin2014learning} tried learning a set of discriminative detectors on the basis of a huge dataset of image patches.

As the deep learning era arrived, approaches focused on exhaustive exploitation of mid-level features from dense local patches~\cite{song2017multi,gong2014multi,wang2017weakly,herranz2016scene} or dense convolutional activations~\cite{yoo2015multi,guo2017locally}, as well as using CNN features extracted from the entire image. 
Other work~\cite{wu2015harvesting,xie2017hybrid} moved away from treating all the regions of an image equally, to address the discriminative ability of local image regions based on deep CNN features. These approaches are built up on a smaller fixed number of bounding boxes for each image, which is more efficient than operating with a huge number of dense patches.
Other approaches~\cite{wang2017knowledge,cheng2018scene} also leveraged co-occurrence of detected object labels for further improvement.
Beyond aggregation of orderless features, some approaches~\cite{izadinia2014incorporating,xu2017scene} further explored the graphical structure of scene based on object detection.
These approaches have been used for some specific applications such as cross-modal scene retrieval~\cite{johnson2015image,qi2017online}.

Our Adi-Red approach extends the related work on discriminative region discovery by using a deep learning classifier-based approach to discover an adaptive number of discriminative regions per image. 
In contrast to existing approaches, our approach does not extract explicit object-level information or use computationally intensive operations such as feature extraction from a huge number of patches or patch clustering.

\subsection{Multi-Scale CNN Features}
Standardly, scene recognition approaches recognize the importance of using features derived from multiple scales.
Although CNN features have shown breakthrough successes in various visual recognition tasks, they have limited geometric invariance to global scaling, which limits their robustness for representing highly variable scenes~\cite{gong2014multi}. 
As a result, most existing deep learning approaches to scene recognition~\cite{song2017multi,wang2017weakly,gong2014multi,wang2017knowledge,herranz2016scene,cheng2018scene,xie2017hybrid,yoo2015multi,guo2017locally,wu2015harvesting} extract deep features independently for multiple scales. With effective encoding methods such as Fisher Vector (FV)~\cite{perronnin2010improving}, Vector of Locally Aggregated Descriptor (VLAD)~\cite{jegou2012aggregating} and Vector of Semantically Aggregated Descriptor (VSAD)~\cite{wang2017knowledge}, these approaches have achieved significant improvement compared with the single image-level representation. 
Since local objects are contained within a larger scene background, it is desirable to combine object-level and scene-level knowledge to arrive at a decision of the scene category for an image. 
Such a combination can be realized by extracting deep features from multi-scale image patches using hybrid CNN models pre-trained on different databases (i.e., ImageNet and Places).

We adopt a multi-scale CNN feature aggregation framework in order to ensure that we are building on the state of the art.
We choose for a basic three-scale framework and implement a simple max pooling method for feature aggregation. 
We anticipate that integrating more scales or designing more sophisticated encoding methods could potentially yield further improvement.
However, here, our focus is not the framework, but rather on the importance of adaptively discovering discriminative local regions for each image.

\subsection{CNN classifier-based discriminative region discovery}
Discriminative regions are important for a range of different types of image analysis. In the area of fine-grained object recognition, recently proposed approaches~\cite{zheng2017learning,he2017fine} have made use of the convolutional responses from a CNN to discover discriminative regions of an image without the need for human-annotated objects or object parts. 
He et al.~\cite{he2017fine} aimed to localize the foreground object as a single region, while Zheng et al.~\cite{zheng2017learning} learned a multi-attention model to localize a fixed number of object parts (i.e., the head, wing, beak, and tail for a bird, and the front/back view, side view, lights, and wheels for a car). 
Our approach makes use of the same general principle. 
However, it is essentially different from these fine-grained object recognition approaches because we address discriminative region discovery for scene recognition, where no foreground/background distinction or well-constrained spatial configurations can be assumed. 
This conceptual difference motivates our insight of selecting an adaptive number of regions per image. 
Further, we use only off-the-shelf CNN features rather than training attention models~\cite{zheng2017learning} from scratch or fine-tuning~\cite{he2017fine}. 
Another aspect is that we not only explore discriminative information for the original database (Places), but also generalize it to a previously unseen dataset (SUN397). 
In contrast, existing use of convolutional responses for discriminative region discovery exploit discriminative information for 
improving solutions on
the same dataset and do not investigate the potential of information transfer.

\section{Approach}
\label{approach}
In this section, we present the whole pipeline of our adaptive discriminative region discovery (Adi-Red) approach. 
The overall objective of our approach is to extract discriminative information from local regions to complement the global-level image representation. 

As shown in Figure 2, the discriminative regions of each input image are discovered by our Adi-Red approach. 
Specifically, the DisNet automatically discovers discriminative regions from each input image by generating the Dis-Map (Section 3.1), and multi-scale patches will be selected based on a simple, yet effective, threshold $T$ (Section 3.2).
The final image representation will be obtained by using a conventional multi-scale feature aggregation framework.

\subsection{Automatic Discriminative Region Discovery}

We adopt the classifier that we use for discriminative region discovery from the literature on weakly-supervised object localization by the CNNs. 
Specifically, a CNN architecture trained with a GAP or global max pooling (GMP) layer is able to localize objects based on only image-level labels~\cite{oquab2015object,zhou2016learning}. 
We extend this idea to automatically discover local discriminative information for scene recognition without making use of any a priori local-level labels (i.e., we use no bounding box or other object-localization ground truth).

Specifically, we choose the GAP-based architecture for our DisNet rather than GMP because GAP promotes the discovery of multiple discriminative image regions since it forces all the activation values in the last convolutional layer to impact the average output score, instead of only the maximum value as with GMP. 
This point is particularly significant for scene recognition because the discriminative visual cues of a scene normally spread in different regions of an image.
In order to exploit discriminative information for hundreds of scene categories, our DisNet is pre-trained on a large-scale scene-centric database (i.e., Places). For each input image, the discriminative map (Dis-Map) is generated using the DisNet to reflect the discriminative ability of different image regions by upsampling the map values into the image size. 

Basically, the Dis-Map is obtained by computing a weighted sum of all the activation maps of the last convolutional layer.
During the training stage, if a filter captures specific patterns for a category, the GAP loss is able to take effect to learn a higher weight for the contribution of this filter to the final prediction of this category. 
As a result, regions of the test image that show a similar pattern that is captured by the filter will contribute more to the final decision to classify the image into this category. 

We formalize the process as follows. Let $w^c_f$ denote the weight corresponding to a scene category $c$ for a filter $f$ in the last convolutional layer (in total, $N$ filters), the predicted probability $S_{out}^c$ output by the softmax layer corresponding to $c$ is defined as:
\begin{equation}
\label{for2}
\begin{aligned}
S_{out}^c&=\sum_{f=1}^N w^c_f \frac{\sum_{x,y=1}^{l} m_f(x, y)}{l^2}\\
&=\frac{1}{l^2} \sum_{x,y=1}^{l} \sum_{f=1}^N w^c_f  m_f(x, y).
\end{aligned}
\end{equation}

Here, $m_f(x, y)$ denotes the activation at location (x, y) of the $l\times l$ activation map produced by filter $f$. For simplicity, the input bias of the softmax is set to 0 since it has little impact on localization performance~\cite{zhou2016learning}. 
Accordingly, the Dis-Map $D_c(x,y)$ could be expressed as:
\begin{equation}
D_c(x,y)=\sum_{f=1}^Nw^c_fm_f(x, y).
\end{equation}
Essentially, $w^c_f$ indicates the importance of filter $f$ for differentiating category $c$ from others. The generated $D_c(x,y)$ with the size of $l\times l$ indicates the importance of the activation at location (x, y) leading to an image classified into category $c$. 
The Dis-Map is created using the scene class prediction of the DisNet.
If ground truth is available, the ground truth can be used instead of the predicted class.

\subsection{Adaptive Region Selection}
In this part, we propose a multi-scale patch selection method to further make use of the information represented by the Dis-Map adaptively per image. 
For the convenience of calculation, the values in the Dis-Map are first normalized into the interval [0,255].

Intuitively, large values in the Dis-Map could indicate that the corresponding regions of the image have strong discriminative power. 
In order to obtain diverse discriminative regions that spread over the image, we search for local maxima of the Dis-Map using a sliding window. 
Specifically, for each $3\times 3$ window with a stride of 1, we select center values that are equal or greater than all of their 8 surrounding locations as the local maxima. 
Local maxima with the same values from two overlapping windows are counted only once to avoid redundant region selection of regions very close to each other. 
Then, the local maxima with values higher than a threshold $T$ are selected as the final discriminative locations. 
We point out that our specific implementation of region selection is not a central feature of Adi-Red. 
We expect that any other conceptually similar implementation would yield a comparable result. 

Finally, we crop square regions centered around each of the final discriminative locations selected. 
For the special case that the resultant patch size is out of the image area, we shift its location to fall within the image. 
We use two different patch sizes (1/4 and 1/16 of the image size) which we refer to as `local scales'.
The sizes are chosen with an eye to capturing complementary local information in the image.

We adopt a three-scale feature aggregation pipeline~\cite{herranz2016scene}.
We extract deep features using CNNs that are pre-trained on the Places (Places-CNN) for the highest (global) scale and the coarse (1/4) local scale, and pre-trained on the ImageNet (ImageNet-CNN) for the fine (1/16) local scale.
At each scale, the input region is resized to fit the required resolution of the corresponding CNN feature extractors. 
Only a simple max pooling (MP) operation is used for intra-scale feature aggregation, 
and the final image representation is obtained by concatenating all the three L2-normalized intra-scale feature vectors. 
We reiterate that our Adi-Red approach is designed on the basis of the discriminative ability of different regions, with the goal of capturing the intuitive properties of scenes by using an adaptive number of discriminative regions for individual images.
Scene classes that need more discriminative regions are automatically represented using more discriminative regions.
At the same time, the use of too many discriminative regions for a scene category is avoided.
This innovation distinguishes our work from other recent region proposal-based approaches~\cite{wu2015harvesting,xie2017hybrid,javed2017object} which fix the number of bounding boxes used for all the images by default, without addressing the properties that distinguish scene classes.

\begin{table}[]
\newcommand{\tabincell}[2]{\begin{tabular}{@{}#1@{}}#2\end{tabular}}
  \caption{Adi-Red compared to the approaches (Dense and Random) using relatively more regions, but no discriminative regions. (Pre-training datasets are reported for the three scales: global/coarse local/fine local; PL= Places and IN=ImageNet)}
  \begin{tabular}{cccccc}
    \toprule
\tabincell{c}{\# of scales\\(global+local)}  &\tabincell{c}{Pre-training\\dataset}&Random &Dense&\tabincell{c}{Adi-Red}\\
    \midrule
   
    1+1 & PL/-/IN& 55.40&59.62&58.99\\
    1+1 & PL/PL/-& 56.88&57.97&59.11\\
    1+2 & PL/PL/IN& 56.90&60.49&\textbf{61.01}\\
  \bottomrule
\end{tabular}
\end{table}

\section{Experiments and Results}
\label{experiments}
In this section, we report the results of experiments demonstrating the performance of our Adi-Red approach and compare it with other existing CNN-based state-of-the-art approaches on the scene recognition benchmark dataset SUN397~\cite{xiao2016sun}. 
Additionally, we carry out a set of analysis experiments on Places365-Standard dataset~\cite{zhou2017places} to provide more detailed insight into the aspects of Adi-Red that make it superior to baselines. 
\subsection{Datasets and Implementation Details}

We test Adi-Red on two datasets. The main experiments are carried out on the SUN397 dataset, because it is a standard scene recognition benchmark and also because it is distinct from the data used to train the classifier-based region detector (i.e., the DisNet). 
The additional experimental analysis is carried out on the Places365-Standard dataset, in order to gain insight into the individual aspects of Adi-Red. We describe the two datasets in turn.

\textbf{SUN397} is a widely used scene-centric database, which includes 397 scene categories with at least 100 images for each category.
We follow the standard evaluation protocol reported in the original paper~\cite{xiao2016sun}, which involves ten fixed splits with 50 images for training and 50 images for testing per category.

\textbf{Places365-Standard} dataset~\cite{zhou2017places} was created by selecting 365 categories that contain more than 4000 images from the large-scale Places database. Our experimental analysis using the Places dataset was conducted on its validation set which includes 100 images per category. We randomly selected 50 images per category for training, and the rest for testing. 

We adopt a ResNet18-based architecture that ends with a GAP layer followed by a softmax layer for our DisNet model. 
Following similar process as in~\cite{zhou2016learning}, the first max-pooling layer is removed for producing a higher-resolution activation map before the GAP layer, resulting in a $14\times14$ Dis-Map. 
This model was pre-trained on the official 1.8 million training images of the Places365-Standard dataset and, as a result, has the ability to automatically learn discriminative information for hundreds of scenes.

We used the multi-scale deep features generated by our approach to train a linear SVM classifier for scene classification with respect to the specific train-test splits described above. 
The parameter $C$ of the SVM classifiers is optimized based on the global-scale features from the training images, and then fixed for all the experiments. 
The value of $C$ used in this paper was 0.02. 
It was chosen with experiments using the Places data, and was confirmed to also work well for the SUN397 data.
We observed a limited sensitivity of the results to the exact value of this parameter.
The classification accuracy, i.e., the percentage of correctly classified images among all the test images, is adopted for evaluation.
Note that accuracy is a standard evaluation metric for scene recognition benchmarks with balanced classes.

\begin{table}
  \caption{Comparison of Adi-Red (multi-scale: global + local) and a baseline (single-scale: global) based on three different CNN architectures on SUN397}
\begin{tabular}{ccc}
    \toprule
    Networks &Baseline & Adi-Red\\
    \midrule
    AlexNet & 54.17& \textbf{61.01}\\
    ResNet18 &66.96 & \textbf{70.58}\\
    ResNet50 &71.38 & \textbf{73.59}\\
 
  \bottomrule
\end{tabular}
\end{table}

\subsection{Scene Recognition Experiments (SUN397)}
In order to evaluate the discriminative information learned by our approach, we carry out extensive experiments on the widely-used scene recognition benchmark SUN397. 
For the training images from the same category as one of the 365 categories of the Places dataset, we use the ground truth label $c$, as used in Eq. (2), for our DisNet. 
For the training images with no overlapping labels (totally 103 categories) with Places or any of the testing images, we instead used their predicted labels decided by the DisNet for generating the Dis-Map.

\subsubsection{Effectiveness of discriminative patches}

In order to demonstrate the effectiveness of discriminative information directly extracted from the DisNet, we compare our Adi-Red approach with two other patch sampling methods (dense sampling and random sampling), which use the same pipeline as Adi-Red, but do not make use of the discriminative ability of local patches. 
Specifically, dense sampling crops 10 patches per image for the middle scale and 50 patches per image for the smallest scale, and random sampling crops 5 random patches per image at each local scale). 
As shown in Table 1, dense sampling method performs best by exhaustively making use of all the fine local patches (ten times the number of patches used by Adi-Red). However, Adi-Red yields better performance for the latter two conditions, confirming the contribution of local discriminative information captured by Adi-Red, and also the benefits of combing scene-level and object-level knowledge for different local scales.
Further, it is important to note that Adi-Red needs only ca. 7 patches per image on average. This leads to much more efficient deep feature extraction, in the feature extraction step, than with the dense sampling method which uses 60 patches per image.

\begin{table*}
\newcommand{\tabincell}[2]{\begin{tabular}{@{}#1@{}}#2\end{tabular}}
\renewcommand\arraystretch{1.2}

 \caption{Adi-Red compared to state-of-the-art scene recognition approaches on the SUN397 dataset, following the standard evaluation protocol. Other approaches use a larger number of scales and number of regions (\# of patches) than Adi-Red.}
 \begin{tabular}{ccccccccc}
    \toprule
    Approach&	Publication	&TS/FT&	\tabincell{c}{\# of scales\\(global+local)}&	dataset&	\tabincell{c}{Encoding\\method}&	\tabincell{c}{\# of\\patches}&		\tabincell{c}{Networks}	&\tabincell{c}{Accuracy\\ (\%)}\\
    \midrule
    MOP-CNN~\cite{gong2014multi}&	ECCV 2014&	-&	1+2	&IN	&VLAD&74&AlexNet&51.98\\
    Context-CNN~\cite{wu2015harvesting}&ICCV 2015&	FT	&0+2	&PL/IN	&VLAD	&256&	AlexNet&	58.11\\
    LS-DHM~\cite{guo2017locally}&	TIP 2017&	FT&	1+0&	PL&	FV&	-&	VGG11&	67.56\\
       Three~\cite{herranz2016scene}&	CVPR 2016	&-&	1+2&	PL/IN	&MP&	$\approx$120	&VGG16&	70.20\\
    Hybrid~\cite{xie2017hybrid}&TCSVT 2017&	-&	1+3&	PL/IN	&MP+FV&	100&\tabincell{c}{VGG19+VGG11+GoogleNet}&70.69\\
    MR-CNN~\cite{wang2017knowledge}&TIP 2017&	TS&	1+2&	PL/IN&	AP	&20&2-resolution BN-inception&72.00\\
MP~\cite{song2017multi}&	TIP 2017	&FT&	1+3&	PL/IN	&SM~\cite{kwitt2012scene}&	276&	VGG16	&72.60\\

Hybrid-PatchNet~\cite{wang2017weakly}&	TIP	2017&TS&	1+9&	PL/IN	&VSAD+FV&1800&BN-Inception+VGG16&	73.00\\
SDO~\cite{cheng2018scene}&PR 2017&FT&	1+1&	PL/IN&	VLAD&	25&VGG16&69.78\\
SDO~\cite{cheng2018scene}&PR 2017&FT&	1+9&	PL/IN&VLAD&$\approx$200&VGG16&73.41\\
Adi-Red&-&-&1+2&PL/IN&MP&$\approx$7&ResNet50&\textbf{73.59}\\
\bottomrule
\end{tabular}
\end{table*}

\subsubsection{Comparison with single-scale global baselines}
In this part, we compare Adi-Red with approaches that use only a single, image-level scale.
We test three different CNN architectures as feature extractors in order to demonstrate that the improvement achieved by Adi-Red is independent of the feature extractor.
As shown in Table 2, our approach outperforms the image-level baseline by a large margin when AlexNet is used. Moreover, although the deeper ResNet models have been equipped with the GAP layer, Adi-Red still achieves better performance. 
These comparisons demonstrate that the discriminative information derived from the DisNet by our approach is complementary to the original global-level CNN features.

\subsubsection{Comparison with the state-of-the-art approaches}
We compare our Adi-Red approach with other scene recognition approaches that have also applied deep CNN features. 
The detailed comparison results are presented in Table 3. 
TS/FT denotes whether the approach involves Training-from-Scratch/Fine-Tuning or not. 
Table 3 reports the number of regions used by each approach (\# of patches) and allows us to observe that all the existing approaches using multi-scale deep features require more patches to complement global-scale image features than Adi-Red.
Most of them pursued further improvement by integrating more local scales or training hybrid models. 
In contrast, Adi-Red achieves new state-of-the-art performance with only 3 scales and about 7 patches per image on average. 
We note that the low number of regions translates into lower computational complexity, and makes Adi-Red suited for efficient scene recognition in practice.

\begin{table}
\newcommand{\tabincell}[2]{\begin{tabular}{@{}#1@{}}#2\end{tabular}}
  \caption{Scene classification accuracy on Places365-Standard validation set with different number of scales based on Places-CNN and/or ImageNet-CNN. (Pre-training datasets are reported for the three scales: global/coarse local/fine local; PL= Places and IN=ImageNet)}
\begin{tabular}{ccc}
    \toprule
    \tabincell{c}{\# of scales\\(global+local)} &\tabincell{c}{Pre-training\\dataset}&Accuracy (\%)\\
    \midrule
    1 & PL& 38.53\\
    1+1 & PL/-/IN& 40.55\\
    1+1 & PL/PL/-& 41.36\\
    1+2 & PL/IN/IN& 39.25\\
    1+2 & PL/PL/PL&41.41\\
    1+2 & PL/PL/IN& \textbf{41.87}\\
  \bottomrule
\end{tabular}
\end{table}

\begin{figure}
\includegraphics[height=0.8\columnwidth,width=\columnwidth]{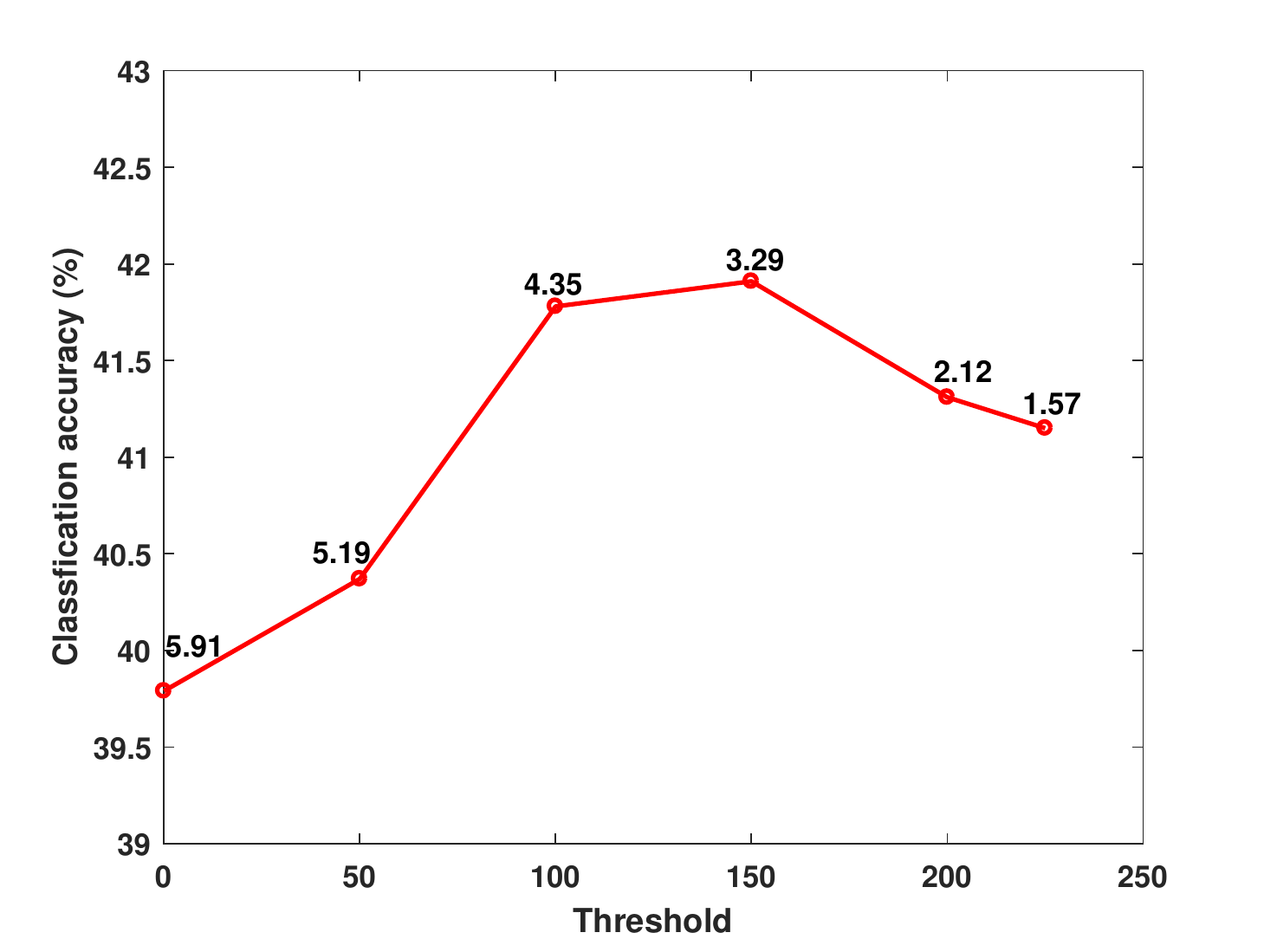}
\caption{Scene classification accuracy on our Places365-Standard test dataset (which is derived from the standard validation set) as a function of the threshold $T$. Numbers above the line denote the average number of regions per image at different threshold levels.}
\end{figure}

\subsection{Additional Experimental Analysis}
In this subsection, we evaluate the effectiveness of individual aspects of Adi-Red using a comparison with other alternative approaches.
For these additional experiments, we evaluate Adi-Red on the Places365-Standard validation set. 
Choosing to test Adi-Red on data from the same domain as the training data of the classifier-based region detector (i.e.,DisNet) eliminates possible impact of domain mismatch on our experimental results, giving us more direct insight into the Adi-Red algorithm.
For fair comparison, the same AlexNet model ~\cite{krizhevsky2012imagenet} will be used as the feature extractor for all the alternative approaches.

\subsubsection{Effectiveness of multi-scale feature aggregation}
We show the effectiveness of our multi-scale feature aggregation pipeline compared with the single-scale baseline in which only global deep features are extracted using the Places-CNN. 
As shown in Table 4, overall, incorporating regions from local and global scales leads to better performance. 
When all three scales are used, using Places-CNN for the global scale and the coarse local scale, and ImageNet-CNN for the fine local scale yields better performance than another two alternatives. 
These results point to the conclusion that Adi-Red is successfully transferring combined knowledge from both scene (Places) and object (ImageNet) databases by applying different deep feature extractors at specific scales.

\begin{figure}
\includegraphics[height=0.8\columnwidth,width=\columnwidth]{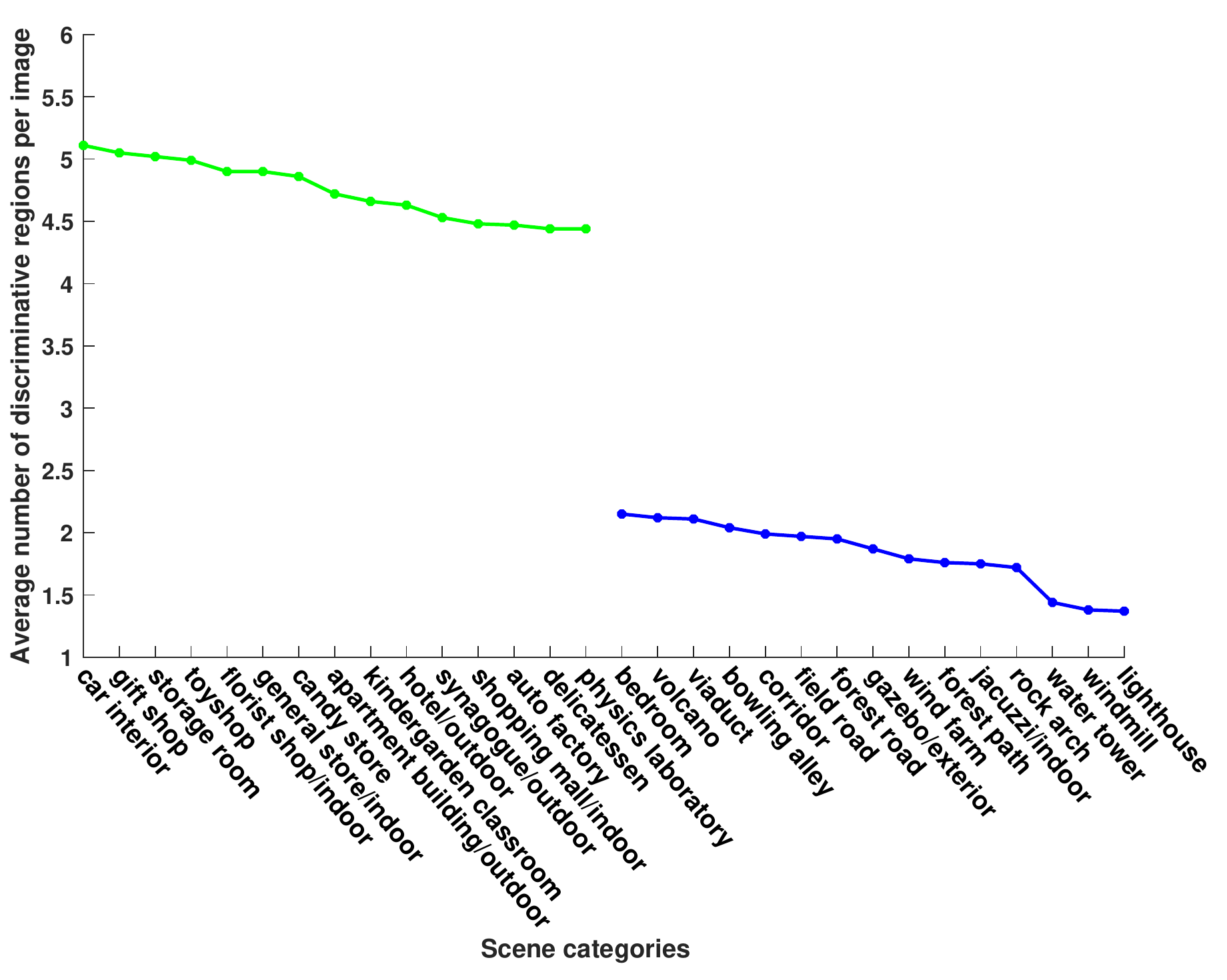}
\caption{
Examples of different scene categories from Places365-Standard validation set with various average number of discriminative regions per image. For readability, we only show the 15 scene categories for which, on average, the most discriminative regions are selected (in green) and another 15 scene categories for which, on average, the fewest discriminative regions are selected (in blue).}
\end{figure}

\subsubsection{Evaluation on Adaptive Region Selection}
We carry out experiments with various values of threshold $T$ to understand the impact of the threshold $T$ on performance.
Figure 3 reports classification accuracy as a function of the threshold. 
The numbers above the line indicate the average number of regions selected per image given a specific threshold. 
Examining the figure reveals that choosing a $T$ that allows for a moderate number of patches per images yields the best performance. 
It also confirms that fewer patches do not provide enough information for good performance.
This observation is consistent with our initial conjecture that using more regions with lower discriminative ability would introduce noisy information. 
In the implementation of this paper, we adopted $T=150$ for the coarse local scale and $T=100$ for the finer scale.
Our motivation for these choices was to avoid redundant selection of content when larger patches overlap and to also allow sufficient discriminative information to be captured by the smaller patches.
Note that the choice of threshold is made by considering the pattern on the Places data that can be observed in Figure 3, and applied in all experiments.
We did not fine-tune the threshold.
In the future, further optimization might yield an additional performance improvement for Adi-Red.

Furthermore, we investigate the performance of Adi-Red on individual scenes.
Figure 4 illustrates the 15 scene categories for which Adi-Red chooses the most discriminative regions and 15 scene categories for which Adi-Red chooses the fewest discriminative regions.
This figure shows that Adi-Red automatically treats scenes like `volcano' differently from scenes like `toyshop'.
Returning to consider the intuitive concept of a scene discussed in Section~\ref{introduction}, this difference is to be expected.
A scene of a `volcano' can be expected to contain a smaller number of distinct local regions than a `toyshop'.
A large portion of visual heterogeneity of volcano images can be treated as noise, whereas the visual heterogeneity of toyshop images provides rich hints of the identity of the scene.
Further, the spatial spread of the discriminative regions in the volcano images is more restricted than in the toyshop images.
Adi-Red behaves consistently with this intuition, discovering a smaller number of discriminative regions for `volcano' (average: 2.20 regions) than for `toyshop' (4.99). 
Figure 5 shows some image examples from these two categories with their discriminative regions. 
We can observe the number of discriminative regions varying image by image and the regions actually capturing discriminative information useful for representing each image. 
Further, it indicates our approach is able to automatically eliminate the characteristic entities, such as person, sky and the buildings, which are not specific enough to the target category. 
\begin{figure}
\includegraphics[height=0.3\textwidth,width=\columnwidth]{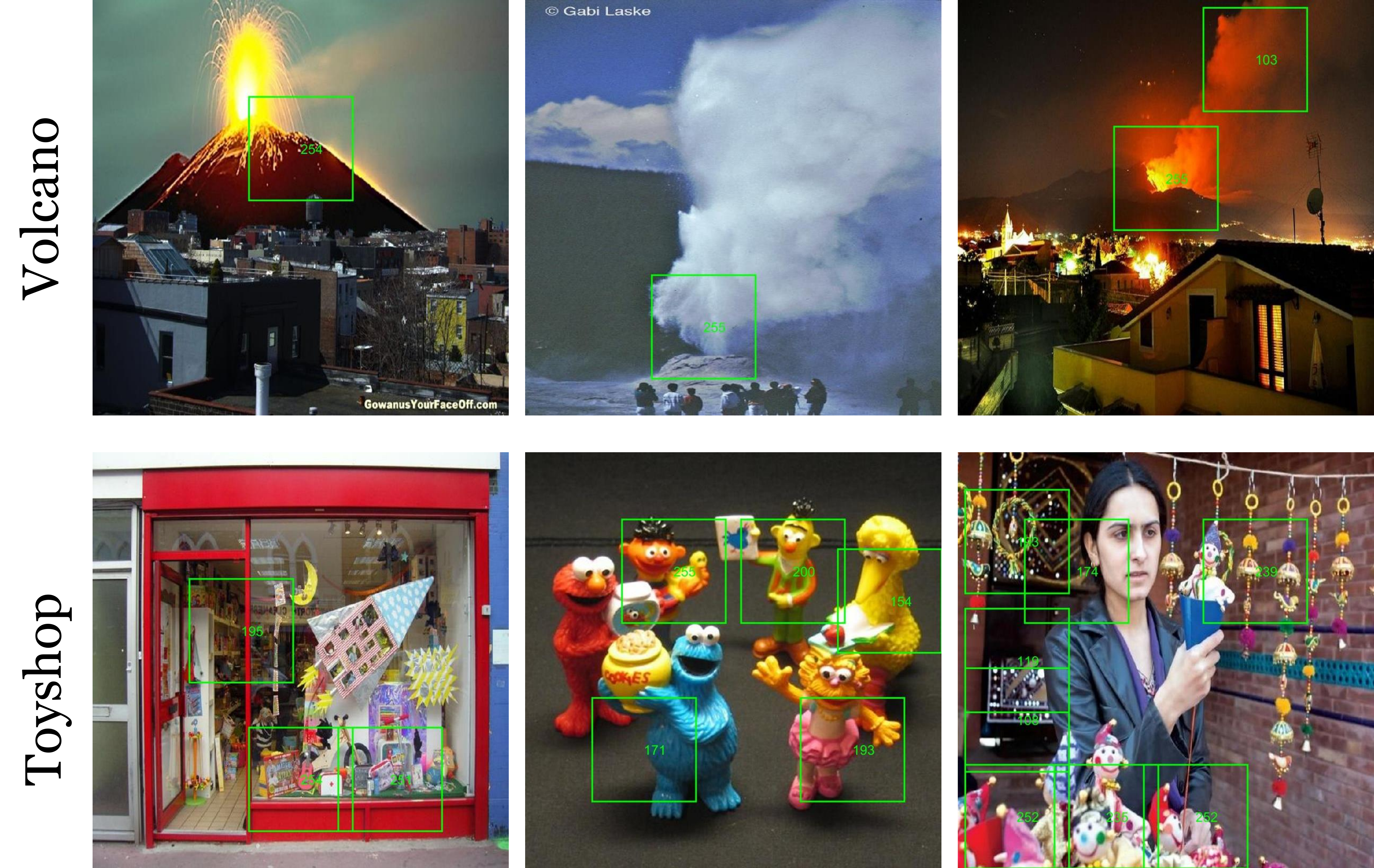}
\caption{Image examples with discriminative regions (marked by green squares with their discriminative ability values) selected by our Adi-Red approach. Only the fine local scale is shown. }
\end{figure}

\begin{figure}
\includegraphics[height=0.45\textwidth,width=\columnwidth]{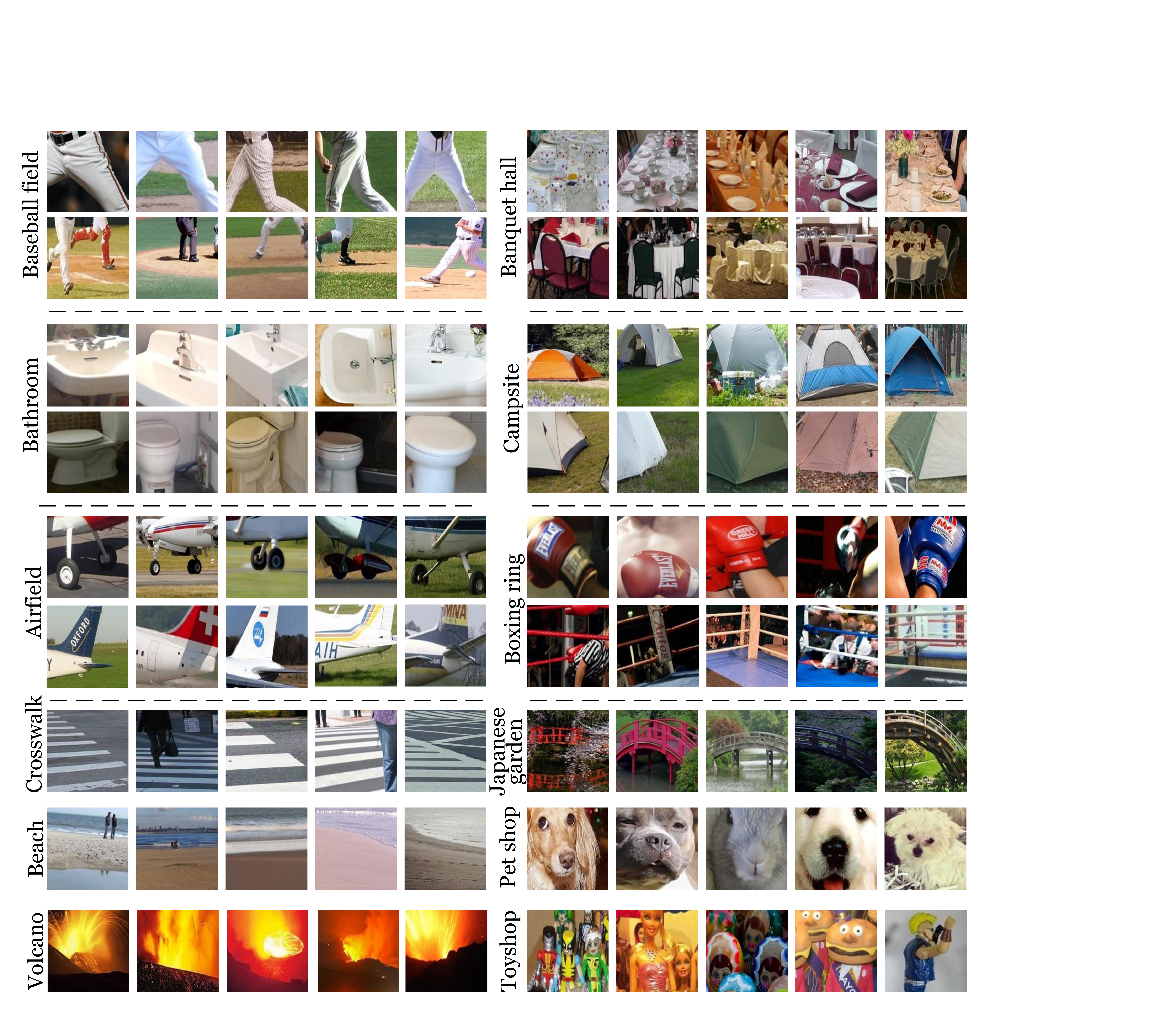}
\caption{Examples of top discriminative patches (fin local scale) selected by our approach for different categories of Places365-Standard validation set.}
\end{figure}

\subsubsection{Visualization of discriminative patches}
In order to demonstrate the discriminative information captured by our Adi-Red approach, in Figure 6, we show some patch examples with highly discriminative ability discovered by our approach on Places365-Standard validation set.
Interestingly, we can see these top discriminative patches capture semantic aspects of different scenes that are also intuitive to human. Specially, we see some patches (from "airfield", "bathroom" and "Japanese garden") convey object-level information, although our approach did not rely on any object priors. 
Other patches capture the interaction between object parts and their surroundings, such as in "baseball field" and "campsite". 
These observations suggest that our approach is discovering semantics related to objects in context. In other words, Adi-Red is going beyond object level to reveal the discriminative pattern of a specific scene. 
In another case, our approach could also capture non-object discriminative information successfully for the scenes (such as "crosswalk" and "beach"), where finding nameable objects (for example, people and cars in the crosswalk scene) based on region proposals is unnecessary and might even introduce confusion instead of contributing to the regions' discriminative properties.

\section{Conclusions}
\label{conclusions}
In this paper, Adi-Red, a novel adaptive discriminative region discovery approach, is proposed for scene recognition.
Adi-Red exploits local discriminative information that is used to complement image-level scene information within a multi-scale deep feature aggregation framework. 
By using a GAP-CNN that has been pre-trained on a large scene-centric dataset (Places), along with a simple threshold-based patch selection strategy, the number of local discriminative regions is allowed to vary adaptively for each image.
Adi-Red is, to our knowledge, the first approach that has used an adaptive number of discriminative regions for the extraction of deep features for scene recognition.
A particularly useful advantage of Adi-Red is that it reduces the computational effort needed to select discriminative patches over existing approaches.

Adi-Red is motivated by an intuitive understanding of what makes scenes special: scenes have a relatively uncontrolled structure (compared with objects in, i.e., fine-grained object detection) and any region of a scene image can provide hints to classify this image into a specific scene category.
These properties inspire us to take an approach that minimizes assumptions about which regions and how many regions will be useful.
Extensive experimental results show the state-of-the-art performance and efficiency achieved by our approach on the scene recognition benchmark dataset SUN397.
Further analysis on individual scene categories reveals that our approach could capture the distinctive properties of diverse scene categories, taking advantage of both object-related information as well as other discriminative patterns not specifically related to objects.
Manual inspection of the results revealed that the semantic content of the discriminative regions that were selected by Adi-Red was sometimes unexpected from the human point of view.
In particular, we identified cases in which the wrong predicted label was used for generating the Dis-Map of the image. 
Future work should focus on how to improve the patch selection method in the case that no ground truth is available for the image, on the basis of retaining useful information without increasing the amount of noise introduced.

We close with a word of outlook. We expect that scene recognition approaches that use multi-scale deep feature frameworks will continue to develop and improve in the coming years. 
The underlying insight of Adi-Red is that scenes should use a varying number of discriminative regions.
In other words, scene recognition should adapt when moving from `volcano' to `toyshop'. Moving forward, we believe that this insight will continue to deliver a performance improvement on top of other improvements gained in scene recognition.

\begin{acks}
The first author, Zhengyu Zhao, thanks the China Scholarship Council for providing monthly allowance during his research towards a doctorate.
\end{acks}

\bibliographystyle{ACM-Reference-Format}
\bibliography{reference_revision}
\end{document}